\documentclass[10pt, a4paper]{article}

\usepackage{lrec-coling2024}
\usepackage{graphicx}
\usepackage{amsmath}
\usepackage{amsfonts}
\usepackage{caption}
\usepackage{float}
\usepackage{subfig}
\usepackage{multirow}

\title{Post-decoder Biasing for End-to-End Speech Recognition of Multi-turn Medical Interview}

\name{Heyang Liu $^1$, Yu Wang$^{1,2}$\sthanks{\quad Corresponding authors}, Yanfeng Wang$^{1,2}$} 

\address{$^{1}$Cooperative Medianet Innovation Center, Shanghai Jiao Tong University \\
  $^{2}$Shanghai Artificial Intelligence Laboratory 
         \\\{liuheyang, yuwangsjtu,wangyanfeng622\}@sjtu.edu.cn\\}

\abstract{
End-to-end (E2E) approach is gradually replacing hybrid models for automatic speech recognition (ASR) tasks. However, the optimization of E2E models lacks an intuitive method for handling decoding shifts, especially in scenarios with a large number of domain-specific rare words that hold specific important meanings. Furthermore, the absence of knowledge-intensive speech datasets in academia has been a significant limiting factor, and the commonly used speech corpora exhibit significant disparities with realistic conversation. To address these challenges, we present Medical Interview (MED-IT), a multi-turn consultation speech dataset that contains a substantial number of knowledge-intensive named entities. We also explore methods to enhance the recognition performance of rare words for E2E models. We propose a novel approach, post-decoder biasing, which constructs a transform probability matrix based on the distribution of training transcriptions. This guides the model to prioritize recognizing words in the biasing list. In our experiments, for subsets of rare words appearing in the training speech between 10 and 20 times, and between 1 and 5 times, the proposed method achieves a relative improvement of 9.3\% and 5.1\%, respectively.
 \\ \newline  }

\begin{document}

\maketitleabstract

\section{Introduction}

Automatic speech recognition (ASR) is a fundamental task that converts speech signals into corresponding textual formats. The hybrid model~\cite{dahl2011context,hinton2012deep} consists of distinct components: an acoustic model, a language model, and a lexicon. Subsequently, the sequence-to-sequence paradigm~\cite{graves2012sequence,sutskever2014sequence,chan2016listen,prabhavalkar2017comparison} has provided an E2E pattern, gradually revealing its potential to disrupt the research field. Although this paradigm has shown great superiority, it often declines in the scenarios of many rare words with low frequency in the training corpus. These rare words often contain important meanings with a significant impact on downstream tasks such as question answering. Contextual automatic speech recognition (CASR)~\cite{aleksic2015bringing,michaely2017keyword,pundak2018deep,alon2019contextual,zhao2019shallow,le2021deep} aims to improve the recognition accuracy of hot words, with the most challenging aspect being rare words. It finds wide-ranging applications, for instance, in medical consultation or company meeting scenarios. The common characteristic is that specialized nouns of a knowledgeable nature hold higher significance, and they may have lower word frequencies during training. We refer to such scenarios as knowledge-intensive contexts, and this paper concentrates on enhancing the performance of E2E models in recognizing rare words within this setting.

The scarcity of speech data in knowledge-intensive scenarios, especially multi-turn medical consultation, has become one of the limiting factors in the academic community. In medical settings, the discourse is replete with medical terminology that is not commonly found in daily language. The significant roles played by these terms are beyond the scope of generic data. The industry possesses the capability to acquire large-scale speech data from specific settings. Google has recorded medical dialogues covering 151 different diseases, totaling 14,000 hours~\cite{chiu2017speech}. However, due to concerns regarding internal corporate information and speaker privacy, such data is not made public. Alternatives adopted by the academic community include training on general datasets or pre-collecting audio from relevant domains, which will lead to severe domain mismatch or great human workload. In an authentic open-source setting, a unified speech dataset rich in named entities serves as a crucial prerequisite for academic methodology comparison and system optimization.

In light of the current shortage of relevant speech corpus, this work focuses on two aspects: we construct an English consultation dataset MED-IT. Subsequently, we conduct research on rare word recognition and propose corresponding algorithms. The main contributions of this paper are as follows:

\textbf{1) Dataset}: We are dedicated to the field of medical consultations and have segmented authentic speech data from four departments to build a knowledge-intensive speech corpus called MED-IT. This corpus consists of multi-turn conversational medical dialogues and includes a substantial number of named entities such as medication names, disease symptoms, and treatment plans. It offers robust data support for CASR experiments and other related research.

\textbf{2) Algorithm}: We propose a novel lightweight and portable scheme called post-decoder biasing to enhance the recognition of rare words. Compared to previous approaches, our method has a minimal impact on the recognition of non-rare words. Furthermore, it can be easily integrated into different E2E models without significant computational cost or decoding latency.

\section{Related Works}

\subsection{Knowledge-Intensive Corpus}

Constructing a speech corpus in the knowledge-intensive scenario presents several challenges. Records in the real world are generally characterized by noisy conditions, which makes it difficult to guarantee speech quality. The introduction of specialized vocabulary implies that the data annotation process requires the involvement of professionals. There are also ethical concerns, such as privacy protection. Prior to data being made public, it necessitates scrutiny and protective measures, especially in contexts like medical consultations. Due to the limited availability of knowledge-intensive speech corpora, most studies on CASR have been conducted using general datasets. Many experiments are conducted on LibriSpeech~\cite{panayotov2015librispeech,le2021deep,han2022improving}, while some researchers prefer GigaSpeech~\cite{chen2021gigaspeech,fox2022improving}.

Knowledge-intensive datasets align more closely with real-world applications, providing excellent material for CASR. To our knowledge, only Earnings-21 \cite{del2021earnings} meets the requirements. It collects English speech data from nine different financial sectors, totaling 39 hours, which includes specialized domain-specific factual terms such as organization, company names, and financial vocabulary. However, there has consistently been a shortage of knowledge-intensive speech data in medical consultation scenarios and many other settings. 

\subsection{E2E Contextual Speech Recognition}

As the superiority of E2E speech recognition systems gradually becomes evident, research in CASR has also shifted towards this paradigm. Previous efforts primarily encompass three approaches: shallow fusion based on language models~\cite{aleksic2015bringing,williams2018contextual,kannan2018analysis,zhao2019shallow}, deep context utilizing attention mechanisms~\cite{pundak2018deep,han2022improving}, and deep biasing based on word pieces~\cite{le2021deep,le2021contextualized,zhang2022end}. Shallow fusion enhances the prediction probability of rare words by refining the language model for specific vocabulary. This approach has been extensively studied in both hybrid models and E2E systems.
The deep context approach explicitly incorporates information from the biasing list into the network architecture. A bias encoder is used to generate embeddings for rare words. Deep biasing takes advantage of the powerful representation and modeling capabilities of neural networks. It constructs a prefix tree at the word pieces level, capturing the concatenation patterns of word segments in the biasing list. This process occurs simultaneously with decoding, facilitating the retrieval of potential subsequent context concatenation patterns.

\subsection{Multimodal Knowledge Fusion}

For multimodal processing, the lexicon of text modality provides an intuitional method for knowledge fusion. Taking the field of multimodal emotion detection as an example, emotion lexicons compass the importance of tokens in the specific task. Words imbued with significant emotional valence are assigned higher weights, consequently affording them greater prominence in subsequent recognition and detection processes. The lexicon has been widely used by concept and knowledge retrieval for enhancing meaningful words~\cite{zhong2019knowledge}. Other works directly fuse the information from the word list with the feature vectors of the textual modality~\cite{zhao2023knowledge}. In the field of speech recognition, a similar approach is proposed by Das et al.~\cite{das2022listen}, which introduces a knowledge graph as an external knowledge base. After the first decoding is completed and one of the hot words is correctly recognized, they guide the model to recognize another real word located at a neighboring node by adding additional language model scores. However, such a knowledge-based lexicon is not always available for CASR. Instead, the biasing list provides the importance rate, which can be fused directly into the decoding process alongside the connection distribution of recognition units obtained from the training transcriptions.

\begin{figure*}[htbp]
    \centering
    \includegraphics[width=1\linewidth]{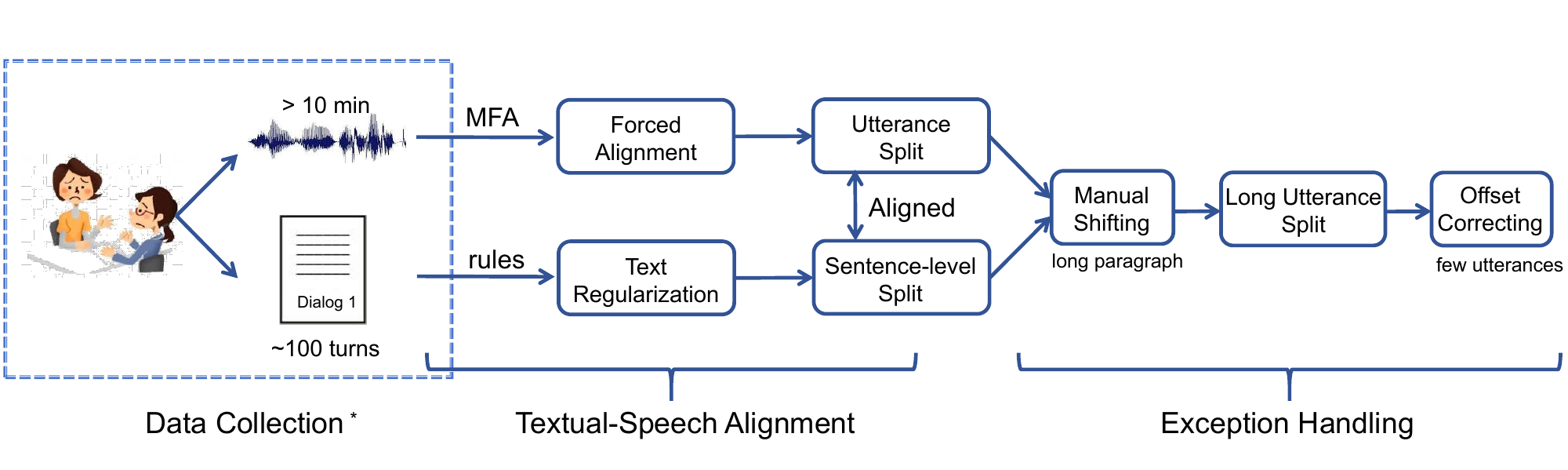}
    \caption{Creation pipeline of MED-IT. Data collection was done in published research. Cleaning and segmentation on both modalities have been performed serialized for textual-speech alignment, and then manual examination for exception handling.}
    \label{fig:creation_pipeline}
\end{figure*}

\section{MED-IT Dataset}

\subsection{Creation Pipeline}

A previous study has collected real-life simulated clinical consultation speech recordings in a hospital ~\cite{fareez2022dataset}. Each dialogue involves one doctor and one patient, both portrayed by medical students, ensuring no potential issues related to privacy disclosure. However, accurate time annotation has not been released, and transcription quality is not trustful. Building upon this open-source corpus, we establish the MED-IT speech dataset, which can be applied to recognition tasks. Our data processing procedure is present in Figure~\ref{fig:creation_pipeline}:


 \textbf{1) Data Cleaning}: We employ a standardized annotation approach for cleaning textual annotation, including using a special token to represent colloquial pauses, converting special characters based on their pronunciation, and systematically rectifying annotation format errors. Speech wav with great noise or overlap is discarded.


 \textbf{2) Data Segmentation}: The original audio segments typically have durations of over ten minutes. Obtaining precise utterance-level alignment is a necessary step in speech recognition training. The Montreal Forced Aligner (MFA)~\cite{mcauliffe2017montreal} is initially applied for forced alignment. We acquire the pronunciation time annotations of each word in a long conversation as a basis for segmenting speech. Sentence-level split on text is performed to ensure the correspondence between speech wav and transcription. Practical experience has shown that the performance of the MFA is closely tied to the acoustic environment. For the few instances where significant alignment errors occur, we utilize the phonetic software Praat~\cite{boersma2001speak} for re-segmentation and annotation. To ensure computational efficiency, we limit the duration of each individual utterance to within 24 seconds. Any statements exceeding this duration are further split.

 \textbf{3) Data Examination}: After the initial segmentation, we conduct regular interval sampling checks on the speech data to prevent any potential audio offsets in shorter passages. Additionally, a small portion of multi-channel audio was converted into single-channel through linear interpolation. The inspected speech data exhibits appropriate durations and precise alignment with the text annotations.

\subsection{Dataset Details and Application}

\begin{figure*}[!t]
\centering

\begin{minipage}[b]{0.4\linewidth}
  \centering
  \subfloat[Distribution of speech duration]{
    \includegraphics[width=\textwidth]{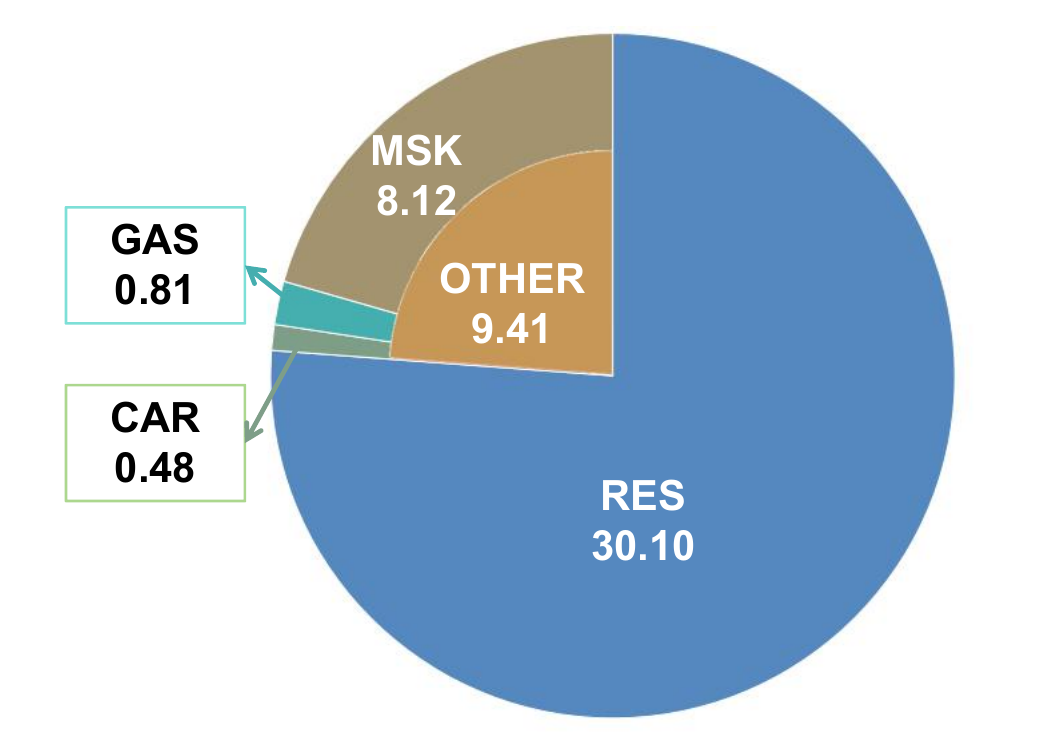}
    \label{subfig:lt}
  }
\end{minipage}
\hfill
\begin{minipage}[b]{0.5\linewidth}
  \centering
  \subfloat[Number of seconds per utterance]{
    \includegraphics[width=\textwidth]{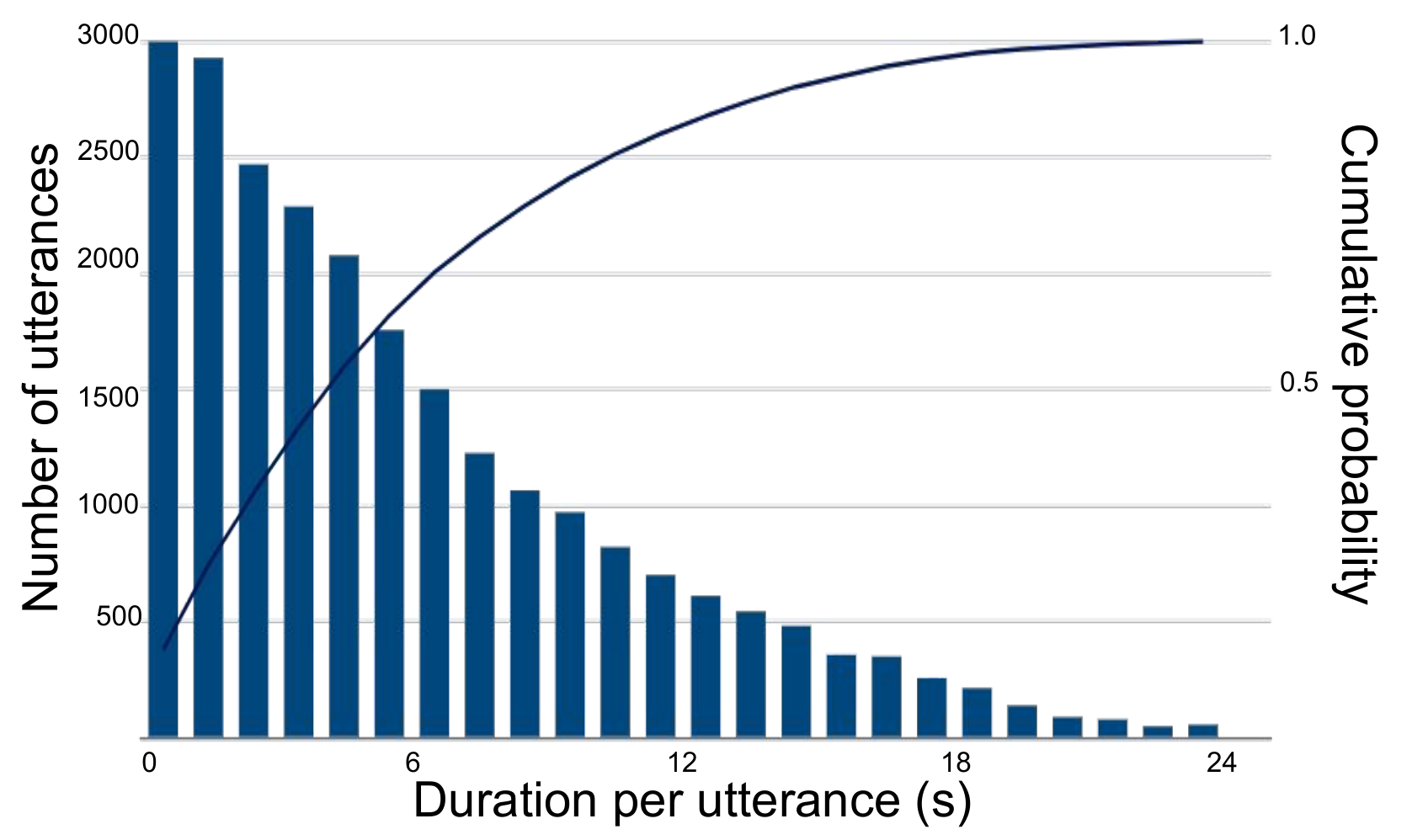}
    \label{subfig:rt}
  }
\end{minipage}

\begin{minipage}[b]{0.42\linewidth}
  \centering
  \subfloat[Partition of different rare words subset]{
    \includegraphics[width=\textwidth]{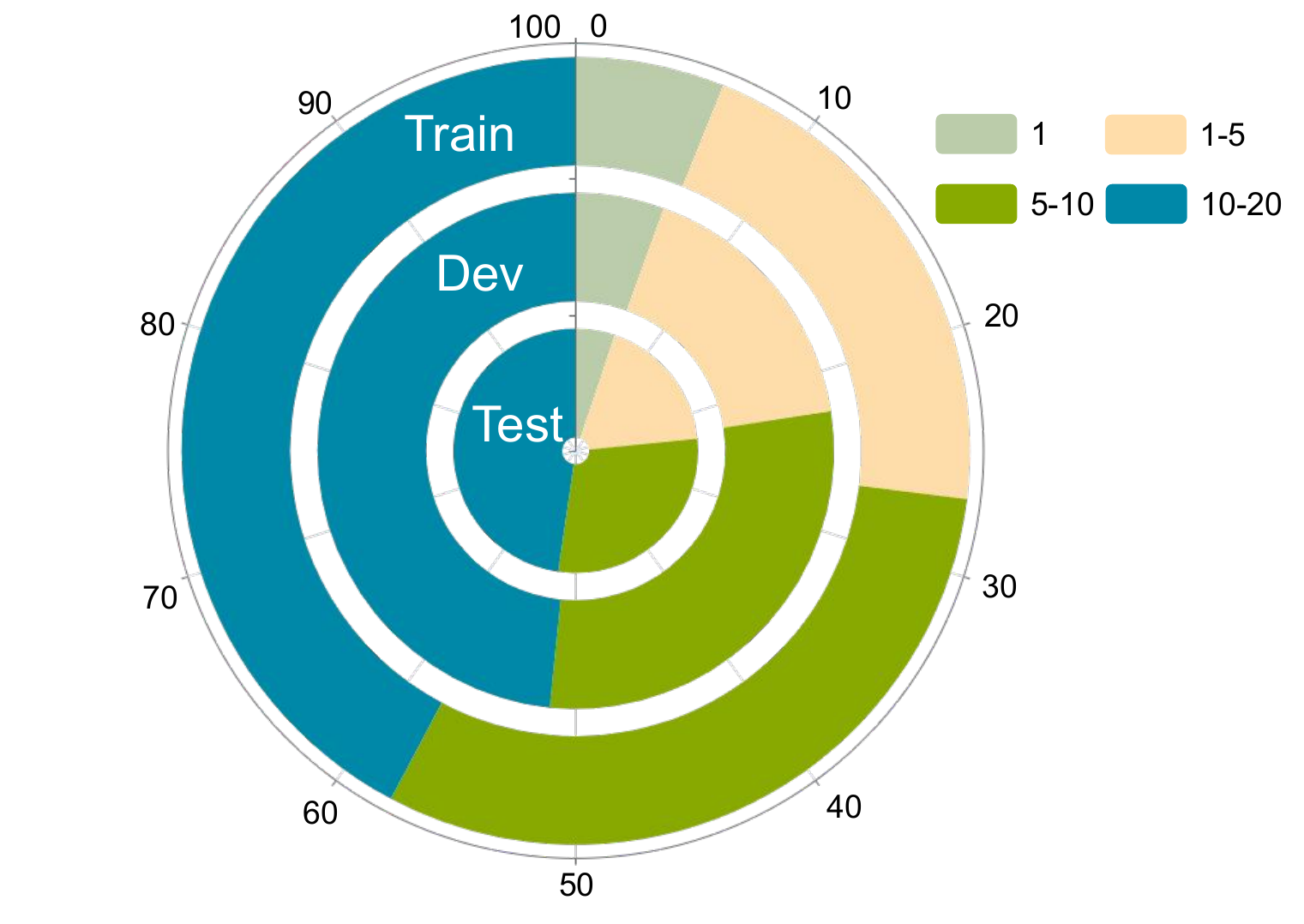}
    \label{subfig:lb}
  }
\end{minipage}
\hfill
\begin{minipage}[b]{0.45\linewidth}
  \centering
  \subfloat[Rare words slices with different frequencies]{
    \includegraphics[width=\textwidth]{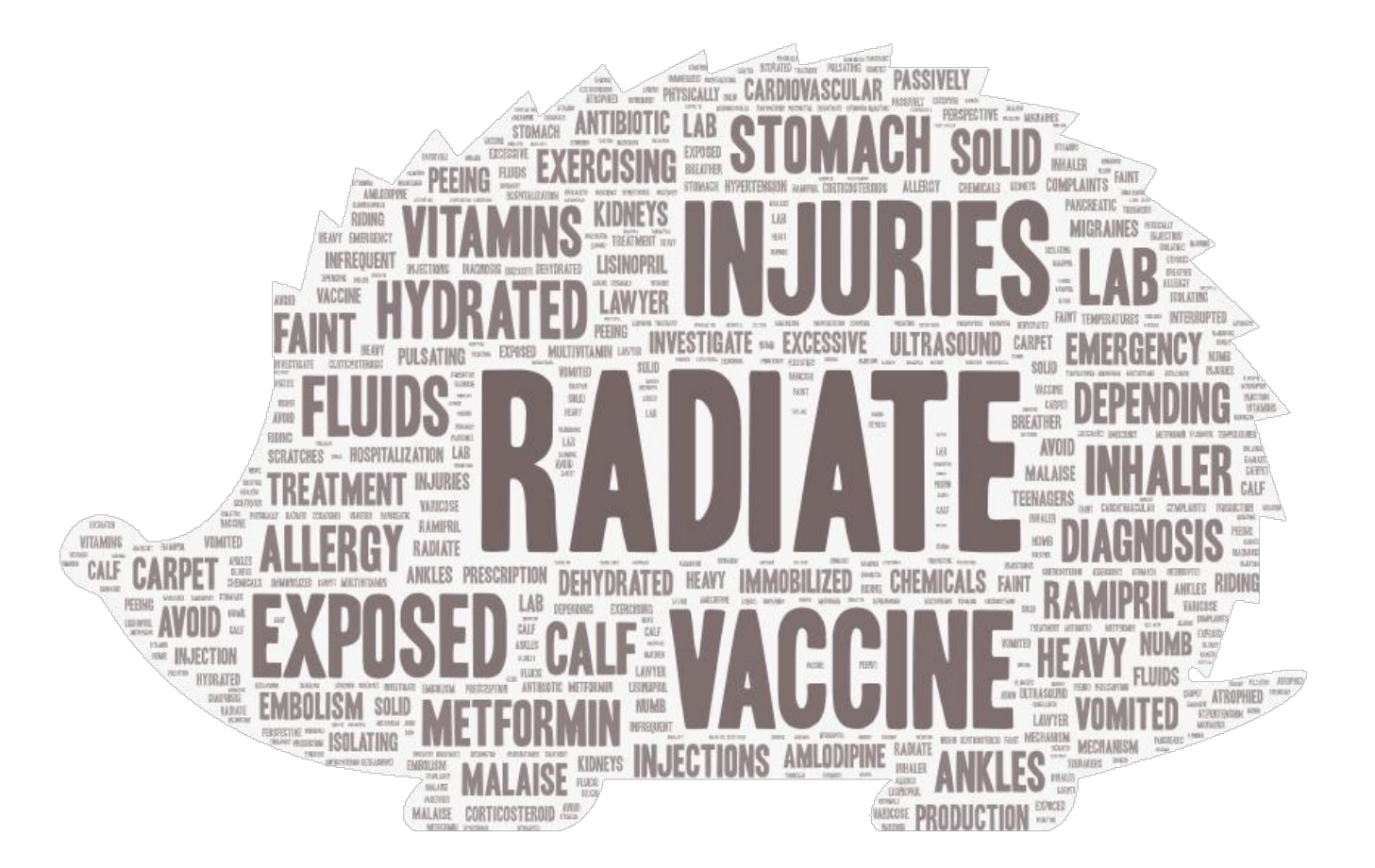}
    \label{subfig:rb}
  }
\end{minipage}

\caption{Statistics of our dataset. (a) shows the speech duration of each department with RES making the most. (b) shows the number of seconds per utterance, with most lasting for less than 10 seconds. (c) indicates the partition portion of each biasing list, and (d) shows word slices from them.}
\label{fig:statistics}
\end{figure*}

MED-IT is a medical consultation speech corpus recorded in real-world scenarios. The doctor-patient dialogues are structured according to the Objective Structured Clinical Examination (OSCE)~\cite{zayyan2011objective}. The main process includes the patient's introduction of symptoms, the doctor's inquiries about the disease condition, and finally, diagnostic and treatment recommendations. Our speech dataset encompasses diagnostic and treatment consultant recordings from four departments: Respiratory (RES), Musculoskeletal (MSK), Gastrointestinal (GAS), and Cardiovascular (CAR). Among these, the majority of the recordings pertain to RES, with the composition ratios and other details as illustrated in Figure~\ref{fig:statistics}.

MED-IT is a standard speech dataset that can be used to evaluate speech recognition models. It contains a significant number of medical-specific terms that are not commonly found in general corpora. This presents a greater and unique challenge for ASR and can be used as evaluation data to assess the generalization capability. The main objective of constructing this dataset is CASR. In addition, MED-IT can also be utilized for research in the field of natural language processing. For example, the manually annotated textual data can be used to train named entity recognition (NER) models with additional annotations. It can also be used to develop semantic understanding and dialogue-generation models for doctor-patient interactions based on the standard OSCE diagnostic process. 

Previous work has shown significant interest in this corpus for ASR tasks, although there is still a performance gap ~\cite{liu2023improved,whetten2023evaluating}. The entire MED-IT dataset and also the annotations have been uploaded to Hugging Face\footnote{The dataset with transcriptions will be made available at \url{https://huggingface.co/datasets/SandO114/Medical_Interview}}.

\subsection{Biasing List Selection for CASR}

As for the CASR task, we select words with training set frequencies falling into the ranges of (10,20], (5,10], (1,5], and 1 to form different subsets of rare words. Figure~\ref{fig:statistics}(c) exhibits the statics of the partition of rare word subsets with different frequencies, with the outer ring indicating the portion of unique words in each rare word set (i.e. word "vaccine" accounts for 1 even if it occurs 20 times), and other rings indicating the total frequency in each evaluation set(i.e. "vaccine" accounts for 5 if it occurs 5 times in corresponding set). Figure~\ref{fig:statistics}(d) shows part of the biasing rare words, with the word size proportional to the relative word frequencies. It can be observed that, although the construction of rare word subsets is based on word frequency rather than semantics, it contains a large number of knowledge-intensive named entities, such as human organs, disease symptoms, and drug names. Rare words carry rich semantic information, and accurately recognizing these vocabulary items will have a profound impact.

\begin{figure*}[!t]
\centering

\subfloat[Caculation of the transform matrix]{
\hspace{-0.4cm}\includegraphics[scale=0.32]{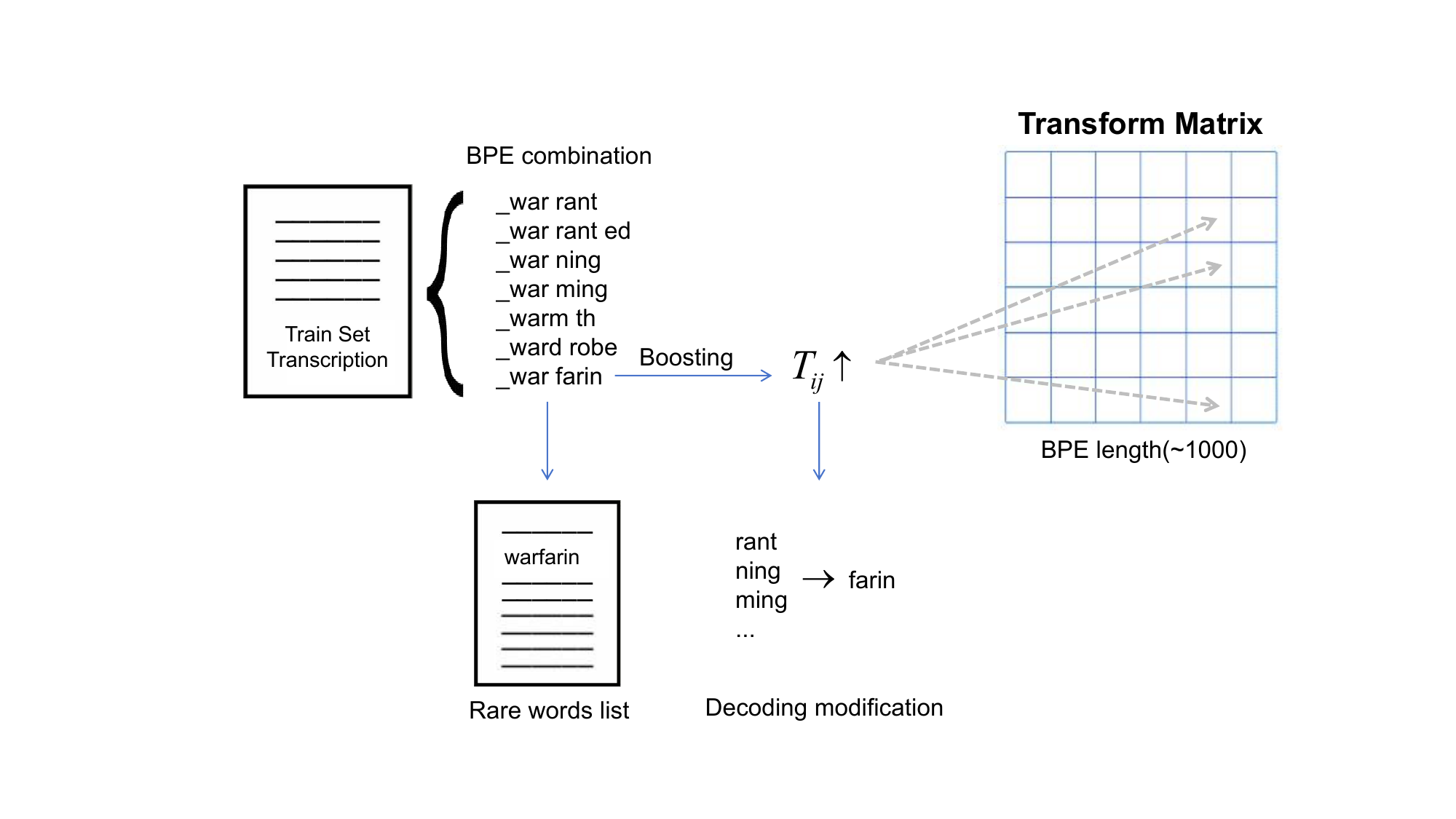}}
\subfloat[LAS model with post-decoder biasing]{
		\includegraphics[scale=0.3]{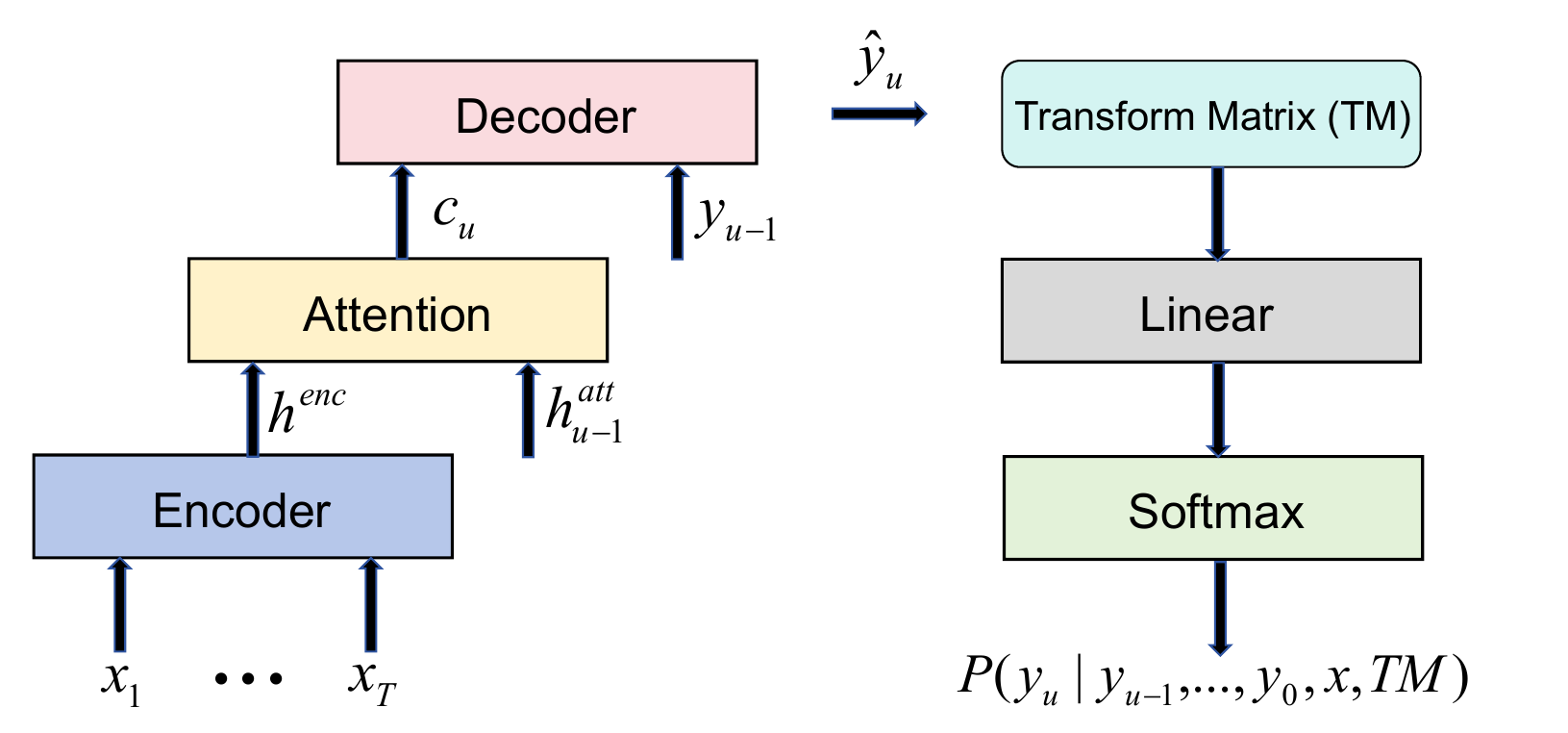}}
\caption{Overview of the post-decoder biasing for attention-based encoder-decoder. The transcription of the train set contains the biasing word "warfarin". BPE "rant" might be transformed to "farin" if the corresponding probability is abnormally high, for "warrant" and "warfarin" are both valid words. The biasing decoding results are determined by both neural architecture and transform matrix obtained from the in-domain sub-word combination distribution.}
\label{fig:postdecoder_biasing}
\end{figure*}

\section{Post-Decoder Biasing}

Rare words can be considered as hot words that appear less frequently in the training set. While the textual transcription in speech recognition inference is unknown, the decoded hypothesis provides an unbiased recognition probability, along with an approximate confidence score. In addition, training transcription provides pivotal connection distribution in-domain, which serves as a steady knowledge base for decoded hypothesis biasing. As illustrated in Figure~\ref{fig:postdecoder_biasing}, the post-decoder fuses the information from the transcription distribution and conducts biased recognition results based on unbiased decoding outputs. This is achieved by the replacement of recognition units that favor the biasing words. The replacement rules are related to the word frequency in the training set. By boosting the word frequencies of rare words, post-decoder biasing guides the model to consider the rare words with a suboptimal recognition probability. 

Compared to previous approaches, our method leverages information from the initial decoding and does not require the introduction of additional knowledge. Furthermore, it avoids the computational cost and latency associated with architectural improvements or secondary decoding fusion.

\subsection{Post-Decoder Architecture}

The post-decoder introduces a transform probability matrix, which facilitates biasing subword replacement. Generally, common words appear more frequently in the training set, leading the model to output higher probabilities for corresponding subword units during inference. On the other hand, rare words tend to have lower probabilities. If the model outputs several subword units with the highest probabilities that are listed as rare words, it suggests the possibility of the model incorrectly identifying a rare word as a common one. Specifically, the decoder outputs token probability with a shape of $[batch, max\_len, bpe\_size]$, where $batch$ denotes the batch size, $max\_len$ represents the maximum number of subwords in a single sentence, and $bpe\_size$ is the number of tokens chosen by the model. When the subwords contained in the rare word list exhibit higher probabilities, the replacement probability matrix executes substitutes from non-rare words to rare words. We incorporate a linear layer connection and use the output probabilities as the basis for recognizing the E2E model. The added network in the post-decoder biasing architecture is a single connected layer, which results in minimal time delay and computation cost.

\subsection{Transform Matrix Calculation}
The transform matrix models the replacement probabilities of byte pair encoding (BPE) units. We do not employ any criteria related to pronunciation but rather consider the concatenations and word combinations in the training transcriptions to obtain an approximate statistical representation. Assuming a BPE partition $B=(b_1,b_2,...,b_k)$, the replacement probability matrix $T$ ideally represents the probability of replacing $b_i$ with $b_j$ as $T_{ij}$. Constructing the transform matrix involves the computation of the following two steps:

\textbf{1) Computing BPE connection probabilities.} We divide BPEs into two sets: word prefixes or standalone tokens $P$ with each element containing the special marker "\_" and word suffixes or mid-segments $S$ without special symbols. For the former, we calculate the subsequent subword unit distribution for this BPE connection as follows:

\begin{equation}
  p_{i}^j=\frac{n_{ij}}{\sum_k n_{ik}} \hspace{5mm} (b_i\in P)
\end{equation}

$p_{i}^j$ represents the probability of subword unit $b_j$ immediately following word prefix (or standalone token) $b_i$ within the corpus domain and without any prior conditions, and $n_{ij}$ denotes the frequency of subword unit $b_j$ immediately following word prefix $b_i$ in the training set. Similarly, for word suffixes and mid-segments, we calculate the distribution of word pieces connected before them:

\begin{equation}
  p_{i}^j=\frac{n_{ji}}{\sum_k n_{ki}} \hspace{5mm} (b_i\in S)
\end{equation}

\textbf{2) Calculating BPE replacement probabilities.} For subword unit $b_i$, assuming the replacement probability is $p_i$, the probability of not being replaced is $1-p_i$, i.e., $T_{ii}=1-p_i$. We model the replacement probability of BPE as the substitutability of subword units in the training text, following the calculation method as follows:

\begin{equation}
  T_{ij}= p_i \sum_k  p_{i}^k p_{k}^j \hspace{5mm} (i\neq j,i\neq k,j\neq k)
\end{equation}

The calculation of the transform matrix is done independently before the training of the neural network, which means it does not affect the efficiency of the training and inferring process. Additionally, we have not introduced any additional prior knowledge or relied on pronunciation-based similarity. In practical applications, this approach exhibits strong scalability, for providing the text annotations of the training set (the in-domain BPE combination distribution) is always sufficient.

\section{Experiments Setup}

\subsection{Dataset and Evaluation}

Experiments are conducted on MED-IT, which has been explained in Section 3. In the following work, we implement E2E models for the CASR task. Most experiments are conducted for biasing lists with frequencies of (10,20] and (1,5] to represent rare words with relatively higher frequencies and extremely low occurrence in the training corpus.

As for the evaluation metrics, WER (word error rate) is the most popular one for assessing the overall performance of a speech recognition system. We introduce a similar concept into rare word recognition, defining RWER (rare word error rate) as the metric for recognizing rare words. It encompasses deletion errors and substitution errors for rare words, as well as insertion errors where rare words wrongly appear in the decoding hypothesis. 

\subsection{Model Specification}

One noteworthy exemplar of E2E speech recognition is the Attention-based Encoder-Decoder (AED)~\cite{chan2016listen}, which utilizes an attention mechanism to enhance the importance of specific information. Connectionist temporal classification (CTC)~\cite{graves2006connectionist} uses intermediate representation allowing repetitions of labels and blank tokens. The model widely used to leverage both advantages is CTC-Attention~\cite{watanabe2017hybrid}, which we use as our recognition backbone. The implementation is carried out following the instructions of the well-known ESPnet~\cite{watanabe2018espnet} toolkit. We use Conformer~\cite{gulati2020conformer} as our encoder, which is improved from Transformer ~\cite{vaswani2017attention} by capturing both long-distance and local representations. Our Conformer encoder contains 6 blocks, with hidden linear units $d^{ff}=1024$ and attention output size $d^{att}=256$. Attention head $H$ is set to 4 and the front CNN kernel size is 31. As for the decoder, we use the Transformer with 6 blocks ($d^{ff}=1024$,$H=8$). We use 1k BPE as our recognition unit obtained by SentencePiece~\cite{kudo2018sentencepiece}. In addition to the attention training objectives, We use a certain degree of CTC loss function. CTC weight $\lambda$ is 0.3 during both training and infering. SpecAugment~\cite{park2019specaugment} is applied with time mask width $T=40$ and frequency mask with $F=30$. We train our model on two 24GB 3090 RTX GPUs for 80 epochs. The top 10 checkpoints are preserved for model averaging. It is worth mentioning that our model does not employ the common batching method used in ESPnet. Instead, we strive to ensure that the same conversation is assigned to the same worker in data parallelization, which provides a certain optimization for the recognition of rare words~\cite{kim2018dialog}.

\section{Results}

\subsection{CTC-Attention Results}

The most common batch processing modes are numel (num element) and unsorted. Both involve random iterations, with the distinction lying in whether the sampling within each batch is sequential or random. We implement dialog batch, which aims to ensure that utterances in the same dialogue are assigned to the same worker in distributed training. Any discrepancies caused by varying utterance quantities are compensated for by padding at the end. The optimal parameters we use and the experimental results obtained with three different batch processing modes are shown in Table~\ref{tab:batch}. RWER(20) stands for the word error rate of the subset with frequencies between 10 and 20 in the training set, and similarly, RWER(10) quantifies the recognition performance of words with frequencies between 5 and 10. Although the unsorted batch achieves the best overall performance on the whole dataset, the dialog batch shows superiority in rare word recognition, which will be used in the following experimental settings. It is worth mentioning that the recognition performance is much better than previous studies with self-trained or commercial systems on the origin corpus(over 21\% WER)~\cite{liu2023improved,whetten2023evaluating}, thanks to our decent dataset preprocess. 

\begin{table*}
\centering
\setlength{\tabcolsep}{1.5mm}
\caption{Experiment results on MED-IT(\%)}
\label{tab:batch}
\begin{tabular}{c c c c c c c c}
\hline
 Method/Batch & biasing set & $p_i$ & WER  & RWER(20) & RWER(10) & RWER(5) & RWER(1) \\
\hline
Azure\cite{whetten2023evaluating} &- &- & 21.0 &- & -  &- & - \\
IFNT\cite{liu2023improved} &- &- & 22.3 &- & -  &- & - \\
\hline
CTC-Attention& & & & & & &  \\
Numel & - & - & 17.4 & 40.4 & 55.9 & 70.0 & 88.1 \\
Unsorted &- &- & \textbf{15.5} & 38.0 & 48.1 & \textbf{66.8} & 84.1 \\
Dialog &- &- & 16.0 & \textbf{37.8} & \textbf{47.1} & 67.3 & \textbf{82.8} \\
\hline
\multirow{6}*{Dialog} & (1,5]& 0.3 & 16.4 & 38.3 & 50.4 & 66.3 & 82.1 \\ 
 &(1,5]& 0.7 & 16.2 & 35.6 & 46.6 & \textbf{64.1} & 80.1 \\
 &(10,20] & 0.3 & 16.4 & \textbf{34.3} & 46.3 & 67.1 & 80.1 \\
 &(10,20] & 0.7 & 16.2 & 36.1 & 49.6 & 68.6 & 83.4 \\
 &(1,5]& Auto & 16.3 & 35.3 & 48.4 & \textbf{63.9} & 82.1\\
 &(10,20]& Auto & 15.9 & \textbf{34.5} & 49.1 & 66.1 & 82.7 \\
\hline

\end{tabular}
\end{table*}

\subsection{Post-Decoder Biasing Results}

To evaluate the performance of post-decoding biasing for CASR, we increase the frequency of each word in the rare word subset of (10,20] and (5,10] by 100 and calculate the transform probability matrix $T$. Before that, it is necessary to quantify $p_i$. Here, we treat $p_i$ as a hyperparameter based on empirical knowledge. In the next subsection, we will use an automatic and quantified method to specify this value. To some extent, $p_i$ affects the confidence threshold of the recognition probability from the decoder: the larger $p_i$, the higher the confidence threshold, and only the optimal outputs that exceed this confidence threshold will be retained, while outputs with confidence below this threshold will be replaced with biasing tokens towards rare words. In general, recognition units with lower word frequencies in the training set tend to have fewer learned features, resulting in smaller corresponding probabilities. Therefore, we choose larger values of $p_i$ for rare words with lower frequencies and smaller values for rare words with relatively higher frequencies. The experimental results with post-decoder biasing are shown in the third part of Table~\ref{tab:batch}.

The baseline experiment using dialog batch without rare words biasing achieves rare word error rates of 37.8\% and 67.3\% for the two subsets respectively. When using the post-decoder and increasing the word frequency with occurrences in the range of (1,5] by 100 times, the corresponding rare word error rate can be reduced to 34.3\% with a relative improvement of 9.3\%. Similarly, increasing the word frequency with occurrences in the range of (10,20] by 100 times leads to a rare word error rate of 64.1\% with 4.8\% relative improvement. 

\subsection{Automatic Tuning Results}

For a subword unit BPE $b_i$, if it occurs frequently enough in the training set, the decoding probability tends to be stable for many speech features learned by the neural model. In this case, we choose a higher $p_i$ which means a high threshold to keep the output unchanged. Let $n_i$ denote the frequency of $b_i$ in the training transcriptions. The auto-selection process is present in Equation~\ref{linear_p}. We follow the linear interpolation scheme and avoid the need to adjust hyperparameters for different frequency ranges of rare word subsets. The experimental results, shown in the third part of Table~\ref{tab:batch}, demonstrate that the automatic adjustment of replacement probabilities leads to similar performance. It achieves a relative reduction of 5.1\% in RWER on the subset of extremely rare words with occurrences in the range of (1, 5]. Furthermore, it shows minimal degradation in performance on the other subset. It is worth noting that various interpolation schemes have minimal influence on the efficacy of rare word recognition. Therefore, the primary consideration should be the associated computational cost.

\begin{equation}
\label{linear_p}
 p_i=\left\{
\begin{aligned}
&0.9 \hspace{18mm} n_i\geq 1000 \\
&0.9 \frac{n_i}{1000} \hspace{8mm} 100<n_i<1000\\
&0.09 \hspace{18mm} n_i\leq 100
\end{aligned}
\right.
\end{equation}

\subsection{Ablation Experiment}

In order to demonstrate the joint effectiveness of each component of the post-decoder, we conduct ablation experiments, constructing the same structure that solely uses the replacement probability matrix, or adds a linear layer. The same architecture without increasing the training set word frequency is also conducted. The experimental results are shown in Table~\ref{tab:part_postdecoder}.

\begin{table*}
\centering
\caption{Ablation Experiment for Post-Decoder(\%)}
\label{tab:part_postdecoder}
\begin{tabular}{c c c c c c c c c}
\hline
linear&TM&biasing set& $p_i$ &  WER & RWER(20) & RWER(10) & RWER(5) & RWER(1) \\
\hline
- & - & - & - & 16.0 & 37.8 & 47.1 & 67.3 & 82.8 \\
\checkmark & - & - & - & 16.5 & 37.1 & 50.1 & 67.3 & 82.8 \\
- & \checkmark & - & 0.3 & 16.4 & 37.6 & 47.9 & 67.3 & 85.4 \\
- & \checkmark & - & 0.7 & 16.3 & 37.3 & 48.1 & 67.3 & 83.4 \\
- & \checkmark & (1,5] & - & 16.6 & 38.4 & 49.1 & 71.0 & 85.4 \\
- & \checkmark & (10,20] & - & 16.8 & 37.8 & 52.6 & 70.3 & 86.1 \\
\checkmark & \checkmark & - & 0.3 & 16.1 & 37.3 & 46.6 &  67.6 & 82.8 \\
\checkmark & \checkmark & - & 0.7 & 16.3 & 37.3 & 48.6  &  66.6 & 86.8\\
\hline
\checkmark & \checkmark & (1,5]& 0.7 & 16.2 & 35.6 & 46.6 & \textbf{64.1} & 80.1 \\
\checkmark & \checkmark &(10,20] & 0.3 & 16.4 & \textbf{34.3} & 46.3 & 67.1 & 80.1 \\
\hline
\end{tabular}
\end{table*}

Compared to the direct decoding results, using only a linear layer as the post-decoder leads to a slight decline in overall recognition performance (from 16.0\% to 16.5\%), while the recognition performance of rare word subsets remains approximately unchanged. Using only the probability replacement matrix without enhancing the probabilities of rare words also results in a degradation of the overall recognition performance. At the same time, no stable gain in rare word recognition is observed. Enhancing specific subsets and changing the transform matrix alters the model's output distribution, leading to a noticeable decline in both the overall accuracy and rare word recognition performance. When only enhancing the (10,20] rare word subset with the replacement probability matrix, the recognition performance of this subset remains stable (first and sixth row), while the recognition abilities of other subsets decline dramatically. We believe that the enhancement of the probability replacement matrix changes the distribution tendency of the model's output, but its quantification is achieved through the linear layer. We implement the complete post-decoder but do not enhance the training set frequencies (seventh and eighth row). The experiment shows a relatively small impact on the overall recognition performance of the model, decreasing from the baseline of 16.0\% to 16.1\% and 16.3\%. However, there is no significant gain in rare word recognition (less than 1\%). The probability replacement matrix without rare word enhancement has a minor impact on the output distribution. Although it provides the possibility of substituting suboptimal paths, it does not guide the model in a specific direction during inference.

\subsection{Rare Words Enhancing Frequency}

\begin{figure}[htbp]
    \centering
    \includegraphics[width=1.05\linewidth]{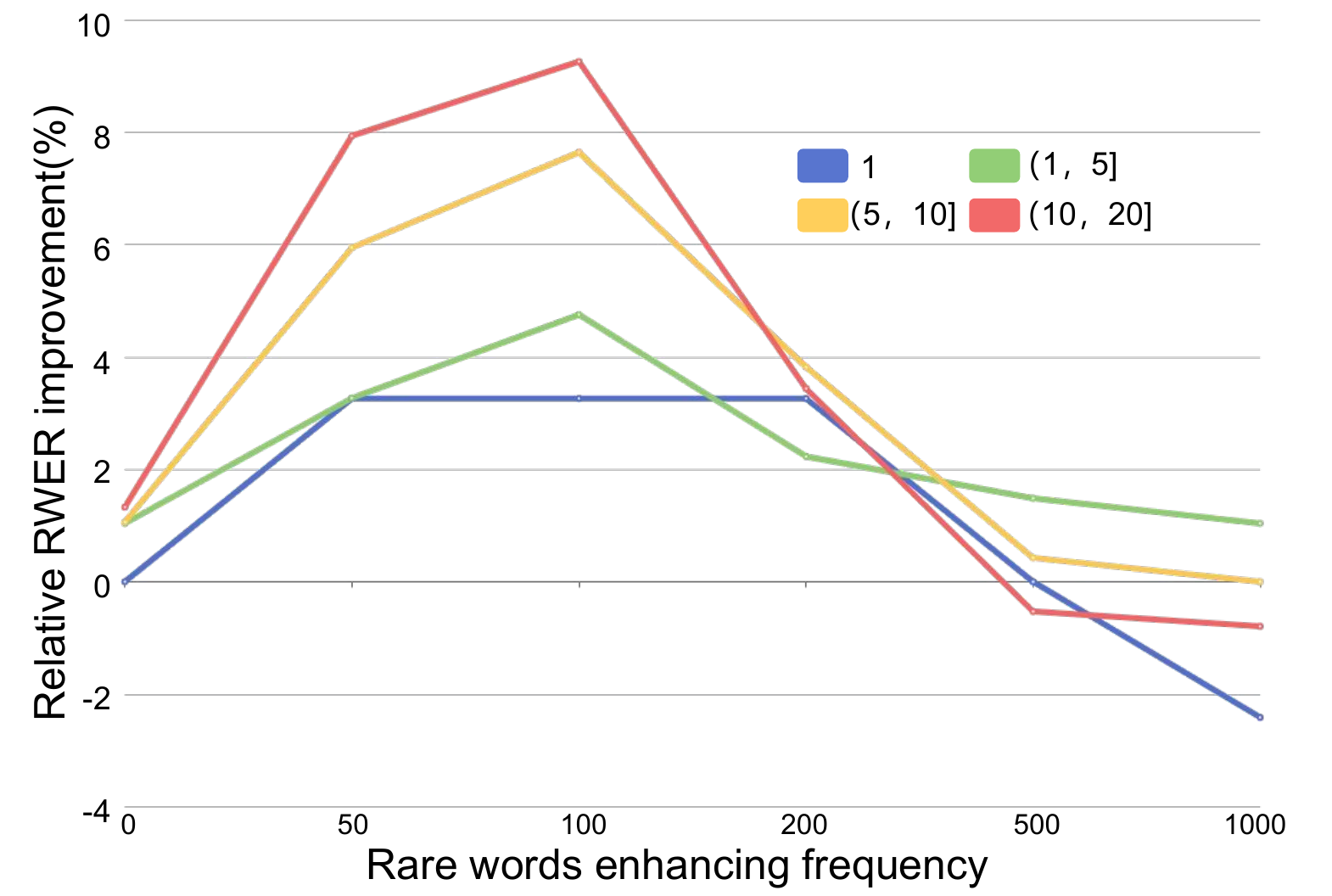}
    \caption{Relative rare words recognition improvement with different enhancing frequencies. Negative values indicate a certain degree of decline compared to the baseline.}
    \label{fig:postdecoder}
\end{figure}

In the previous experiments, we increase the word frequency of the biasing subset in the training set by 100 times when constructing the transform matrix. In this subsection, we investigate the effects of the increased rare word frequency on the recognition performance. We use six different frequencies to construct the replacement probability matrix. The performance of post-decoder biasing with different enhancing frequencies are illustrated in Figure~\ref{fig:postdecoder}.

The experimental results indicate that the frequency enhancement affects the performance of post-decoder biasing. For rare words with extremely low frequencies in the training set, even with an enhancement of 1000 times, there is still a performance gain. However, for other rare word subsets, a certain decrease in recognition performance is observed. As for relatively higher-frequency rare words, when the frequency enhancement exceeds 500 or 1000 times, the post-decoder shows a negative impact on this rare word subset. In general, with higher frequency enhancements, the model's replacement probability increases, leading to a decline in rare word deletion errors and an increase in insertion errors. The increase in insertion errors is more noticeable when the rare word subset itself has a relatively high decoding score. Among the enhancing frequencies, 100 shows the best performance for all the biasing lists. It is encouraging because when the experiment performs well on a specific subset of rare words, this hyperparameter should also be applicable to other subsets. 

\section{Conclusion}

In knowledge-intensive scenarios, rare words often have extremely important meanings. However, the scarcity of speech datasets in this context has limited academic research. In this study, we reconstruct a speech corpus focused on medical inquiries, which contains a wealth of specialized named entities.  To enhance the E2E model's ability to recognize rare words, we propose a lightweight and easily transferable post-decoder biasing method. The experiments show that post-decoder has a positive effect on CASR. By simply increasing the word frequency, the model achieves relative performance improvements of 9.3\% and 5.1\% on two subsets of rare words with frequency ranges of (10,20] and (1,5], respectively. In future work, we will explore rare word replacements based on pronunciation similarities and word-level rules.

\newpage
\nocite{*}
\section{Bibliographical References}\label{sec:reference}
\vspace{-1mm}

\bibliographystyle{lrec-coling2024-natbib}
\bibliography{lrec-coling2024-example}

\begin{thebibliography}{39}
\expandafter\ifx\csname natexlab\endcsname\relax\def\natexlab#1{#1}\fi

\bibitem[{Aleksic et~al.(2015)Aleksic, Ghodsi, Michaely, Allauzen, Hall, Roark, Rybach, and Moreno}]{aleksic2015bringing}
Petar Aleksic, Mohammadreza Ghodsi, Assaf Michaely, Cyril Allauzen, Keith Hall, Brian Roark, David Rybach, and Pedro Moreno. 2015.
\newblock Bringing contextual information to google speech recognition.

\bibitem[{Alon et~al.(2019)Alon, Pundak, and Sainath}]{alon2019contextual}
Uri Alon, Golan Pundak, and Tara~N Sainath. 2019.
\newblock Contextual speech recognition with difficult negative training examples.
\newblock In \emph{ICASSP 2019-2019 IEEE International Conference on Acoustics, Speech and Signal Processing (ICASSP)}, pages 6440--6444. IEEE.

\bibitem[{Boersma and Van~Heuven(2001)}]{boersma2001speak}
Paul Boersma and Vincent Van~Heuven. 2001.
\newblock Speak and unspeak with praat.
\newblock \emph{Glot International}, 5(9/10):341--347.

\bibitem[{Chan et~al.(2016)Chan, Jaitly, Le, and Vinyals}]{chan2016listen}
William Chan, Navdeep Jaitly, Quoc Le, and Oriol Vinyals. 2016.
\newblock Listen, attend and spell: A neural network for large vocabulary conversational speech recognition.
\newblock In \emph{2016 IEEE international conference on acoustics, speech and signal processing (ICASSP)}, pages 4960--4964. IEEE.

\bibitem[{Chen et~al.(2021)Chen, Chai, Wang, Du, Zhang, Weng, Su, Povey, Trmal, Zhang et~al.}]{chen2021gigaspeech}
Guoguo Chen, Shuzhou Chai, Guanbo Wang, Jiayu Du, Wei-Qiang Zhang, Chao Weng, Dan Su, Daniel Povey, Jan Trmal, Junbo Zhang, et~al. 2021.
\newblock Gigaspeech: An evolving, multi-domain asr corpus with 10,000 hours of transcribed audio.
\newblock \emph{arXiv preprint arXiv:2106.06909}.

\bibitem[{Chiu et~al.(2017)Chiu, Tripathi, Chou, Co, Jaitly, Jaunzeikare, Kannan, Nguyen, Sak, Sankar et~al.}]{chiu2017speech}
Chung-Cheng Chiu, Anshuman Tripathi, Katherine Chou, Chris Co, Navdeep Jaitly, Diana Jaunzeikare, Anjuli Kannan, Patrick Nguyen, Hasim Sak, Ananth Sankar, et~al. 2017.
\newblock Speech recognition for medical conversations.
\newblock \emph{arXiv preprint arXiv:1711.07274}.

\bibitem[{Dahl et~al.(2011)Dahl, Yu, Deng, and Acero}]{dahl2011context}
George~E Dahl, Dong Yu, Li~Deng, and Alex Acero. 2011.
\newblock Context-dependent pre-trained deep neural networks for large-vocabulary speech recognition.
\newblock \emph{IEEE Transactions on audio, speech, and language processing}, 20(1):30--42.

\bibitem[{Das et~al.(2022)Das, Sunkara, Bekal, Chau, Bodapati, and Kirchhoff}]{das2022listen}
Nilaksh Das, Monica Sunkara, Dhanush Bekal, Duen~Horng Chau, Sravan Bodapati, and Katrin Kirchhoff. 2022.
\newblock Listen, know and spell: Knowledge-infused subword modeling for improving asr performance of oov named entities.
\newblock In \emph{ICASSP 2022-2022 IEEE International Conference on Acoustics, Speech and Signal Processing (ICASSP)}, pages 7887--7891. IEEE.

\bibitem[{Del~Rio et~al.(2021)Del~Rio, Delworth, Westerman, Huang, Bhandari, Palakapilly, McNamara, Dong, Zelasko, and Jett{\'e}}]{del2021earnings}
Miguel Del~Rio, Natalie Delworth, Ryan Westerman, Michelle Huang, Nishchal Bhandari, Joseph Palakapilly, Quinten McNamara, Joshua Dong, Piotr Zelasko, and Miguel Jett{\'e}. 2021.
\newblock Earnings-21: A practical benchmark for asr in the wild.
\newblock \emph{arXiv preprint arXiv:2104.11348}.

\bibitem[{Fareez et~al.(2022)Fareez, Parikh, Wavell, Shahab, Chevalier, Good, De~Blasi, Rhouma, McMahon, Lam et~al.}]{fareez2022dataset}
Faiha Fareez, Tishya Parikh, Christopher Wavell, Saba Shahab, Meghan Chevalier, Scott Good, Isabella De~Blasi, Rafik Rhouma, Christopher McMahon, Jean-Paul Lam, et~al. 2022.
\newblock A dataset of simulated patient-physician medical interviews with a focus on respiratory cases.
\newblock \emph{Scientific Data}, 9(1):313.

\bibitem[{Fox and Delworth(2022)}]{fox2022improving}
Jennifer~Drexler Fox and Natalie Delworth. 2022.
\newblock Improving contextual recognition of rare words with an alternate spelling prediction model.
\newblock \emph{arXiv preprint arXiv:2209.01250}.

\bibitem[{Graves(2012)}]{graves2012sequence}
Alex Graves. 2012.
\newblock Sequence transduction with recurrent neural networks.
\newblock \emph{arXiv preprint arXiv:1211.3711}.

\bibitem[{Graves et~al.(2006)Graves, Fern{\'a}ndez, Gomez, and Schmidhuber}]{graves2006connectionist}
Alex Graves, Santiago Fern{\'a}ndez, Faustino Gomez, and J{\"u}rgen Schmidhuber. 2006.
\newblock Connectionist temporal classification: labelling unsegmented sequence data with recurrent neural networks.
\newblock In \emph{Proceedings of the 23rd international conference on Machine learning}, pages 369--376.

\bibitem[{Gulati et~al.(2020)Gulati, Qin, Chiu, Parmar, Zhang, Yu, Han, Wang, Zhang, Wu et~al.}]{gulati2020conformer}
Anmol Gulati, James Qin, Chung-Cheng Chiu, Niki Parmar, Yu~Zhang, Jiahui Yu, Wei Han, Shibo Wang, Zhengdong Zhang, Yonghui Wu, et~al. 2020.
\newblock Conformer: Convolution-augmented transformer for speech recognition.
\newblock \emph{Interspeech 2020}.

\bibitem[{Han et~al.(2022)Han, Dong, Liang, Cai, Zhou, Ma, and Xu}]{han2022improving}
Minglun Han, Linhao Dong, Zhenlin Liang, Meng Cai, Shiyu Zhou, Zejun Ma, and Bo~Xu. 2022.
\newblock Improving end-to-end contextual speech recognition with fine-grained contextual knowledge selection.
\newblock In \emph{ICASSP 2022-2022 IEEE International Conference on Acoustics, Speech and Signal Processing (ICASSP)}, pages 8532--8536. IEEE.

\bibitem[{Hinton et~al.(2012)Hinton, Deng, Yu, Dahl, Mohamed, Jaitly, Senior, Vanhoucke, Nguyen, Sainath et~al.}]{hinton2012deep}
Geoffrey Hinton, Li~Deng, Dong Yu, George~E Dahl, Abdel-rahman Mohamed, Navdeep Jaitly, Andrew Senior, Vincent Vanhoucke, Patrick Nguyen, Tara~N Sainath, et~al. 2012.
\newblock Deep neural networks for acoustic modeling in speech recognition: The shared views of four research groups.
\newblock \emph{IEEE Signal processing magazine}, 29(6):82--97.

\bibitem[{Kannan et~al.(2018)Kannan, Wu, Nguyen, Sainath, Chen, and Prabhavalkar}]{kannan2018analysis}
Anjuli Kannan, Yonghui Wu, Patrick Nguyen, Tara~N Sainath, Zhijeng Chen, and Rohit Prabhavalkar. 2018.
\newblock An analysis of incorporating an external language model into a sequence-to-sequence model.
\newblock In \emph{2018 IEEE International Conference on Acoustics, Speech and Signal Processing (ICASSP)}, pages 1--5828. IEEE.

\bibitem[{Kim and Metze(2018)}]{kim2018dialog}
Suyoun Kim and Florian Metze. 2018.
\newblock Dialog-context aware end-to-end speech recognition.
\newblock In \emph{2018 IEEE Spoken Language Technology Workshop (SLT)}, pages 434--440. IEEE.

\bibitem[{Kudo and Richardson(2018)}]{kudo2018sentencepiece}
Taku Kudo and John Richardson. 2018.
\newblock Sentencepiece: A simple and language independent subword tokenizer and detokenizer for neural text processing.
\newblock \emph{EMNLP 2018}, page~66.

\bibitem[{Le et~al.(2021{\natexlab{a}})Le, Jain, Keren, Kim, Shi, Mahadeokar, Chan, Shangguan, Fuegen, Kalinli et~al.}]{le2021contextualized}
Duc Le, Mahaveer Jain, Gil Keren, Suyoun Kim, Yangyang Shi, Jay Mahadeokar, Julian Chan, Yuan Shangguan, Christian Fuegen, Ozlem Kalinli, et~al. 2021{\natexlab{a}}.
\newblock Contextualized streaming end-to-end speech recognition with trie-based deep biasing and shallow fusion.
\newblock \emph{arXiv preprint arXiv:2104.02194}.

\bibitem[{Le et~al.(2021{\natexlab{b}})Le, Keren, Chan, Mahadeokar, Fuegen, and Seltzer}]{le2021deep}
Duc Le, Gil Keren, Julian Chan, Jay Mahadeokar, Christian Fuegen, and Michael~L Seltzer. 2021{\natexlab{b}}.
\newblock Deep shallow fusion for rnn-t personalization.
\newblock In \emph{2021 IEEE Spoken Language Technology Workshop (SLT)}, pages 251--257. IEEE.

\bibitem[{Liu et~al.(2023)Liu, Yu, and Chen}]{liu2023improved}
Junzhe Liu, Jianwei Yu, and Xie Chen. 2023.
\newblock Improved factorized neural transducer model for text-only domain adaptation.
\newblock \emph{arXiv preprint arXiv:2309.09524}.

\bibitem[{McAuliffe et~al.(2017)McAuliffe, Socolof, Mihuc, Wagner, and Sonderegger}]{mcauliffe2017montreal}
Michael McAuliffe, Michaela Socolof, Sarah Mihuc, Michael Wagner, and Morgan Sonderegger. 2017.
\newblock Montreal forced aligner: Trainable text-speech alignment using kaldi.
\newblock In \emph{Interspeech}, volume 2017, pages 498--502.

\bibitem[{Michaely et~al.(2017)Michaely, Zhang, Simko, Parada, and Aleksic}]{michaely2017keyword}
Assaf~Hurwitz Michaely, Xuedong Zhang, Gabor Simko, Carolina Parada, and Petar Aleksic. 2017.
\newblock Keyword spotting for google assistant using contextual speech recognition.
\newblock In \emph{2017 IEEE Automatic Speech Recognition and Understanding Workshop (ASRU)}, pages 272--278. IEEE.

\bibitem[{Panayotov et~al.()Panayotov, Chen, Povey, and Khudanpur}]{panayotov2015librispeech}
Vassil Panayotov, Guoguo Chen, Daniel Povey, and Sanjeev Khudanpur.
\newblock Librispeech: An asr corpus based on public domain audio books.
\newblock In \emph{2015 IEEE International Conference on Acoustics, Speech and Signal Processing (ICASSP)}.

\bibitem[{Park et~al.(2019)Park, Chan, Zhang, Chiu, Zoph, Cubuk, and Le}]{park2019specaugment}
Daniel~S Park, William Chan, Yu~Zhang, Chung-Cheng Chiu, Barret Zoph, Ekin~D Cubuk, and Quoc~V Le. 2019.
\newblock Specaugment: A simple data augmentation method for automatic speech recognition.
\newblock \emph{Interspeech 2019}.

\bibitem[{Prabhavalkar et~al.(2017)Prabhavalkar, Rao, Sainath, Li, Johnson, and Jaitly}]{prabhavalkar2017comparison}
Rohit Prabhavalkar, Kanishka Rao, Tara~N Sainath, Bo~Li, Leif Johnson, and Navdeep Jaitly. 2017.
\newblock A comparison of sequence-to-sequence models for speech recognition.

\bibitem[{Pundak et~al.(2018)Pundak, Sainath, Prabhavalkar, Kannan, and Zhao}]{pundak2018deep}
Golan Pundak, Tara~N Sainath, Rohit Prabhavalkar, Anjuli Kannan, and Ding Zhao. 2018.
\newblock Deep context: end-to-end contextual speech recognition.
\newblock In \emph{2018 IEEE spoken language technology workshop (SLT)}, pages 418--425. IEEE.

\bibitem[{Sutskever et~al.(2014)Sutskever, Vinyals, and Le}]{sutskever2014sequence}
Ilya Sutskever, Oriol Vinyals, and Quoc~V Le. 2014.
\newblock Sequence to sequence learning with neural networks.
\newblock \emph{Advances in neural information processing systems}, 27.

\bibitem[{Vaswani et~al.(2017)Vaswani, Shazeer, Parmar, Uszkoreit, Jones, Gomez, Kaiser, and Polosukhin}]{vaswani2017attention}
Ashish Vaswani, Noam Shazeer, Niki Parmar, Jakob Uszkoreit, Llion Jones, Aidan~N Gomez, {\L}ukasz Kaiser, and Illia Polosukhin. 2017.
\newblock Attention is all you need.
\newblock \emph{Advances in neural information processing systems}, 30.

\bibitem[{Watanabe et~al.(2018)Watanabe, Hori, Karita, Hayashi, Nishitoba, Unno, Enrique Yalta~Soplin, Heymann, Wiesner, Chen et~al.}]{watanabe2018espnet}
Shinji Watanabe, Takaaki Hori, Shigeki Karita, Tomoki Hayashi, Jiro Nishitoba, Yuya Unno, Nelson Enrique Yalta~Soplin, Jahn Heymann, Matthew Wiesner, Nanxin Chen, et~al. 2018.
\newblock Espnet: End-to-end speech processing toolkit.
\newblock \emph{INTERSPEECH 2018, Hyderabad, India}.

\bibitem[{Watanabe et~al.(2017)Watanabe, Hori, Kim, Hershey, and Hayashi}]{watanabe2017hybrid}
Shinji Watanabe, Takaaki Hori, Suyoun Kim, John~R Hershey, and Tomoki Hayashi. 2017.
\newblock Hybrid ctc/attention architecture for end-to-end speech recognition.
\newblock \emph{IEEE Journal of Selected Topics in Signal Processing}, 11(8):1240--1253.

\bibitem[{Whetten et~al.(2023)Whetten, Imtiaz, and Kennington}]{whetten2023evaluating}
Ryan Whetten, Mir~Tahsin Imtiaz, and Casey Kennington. 2023.
\newblock Evaluating automatic speech recognition in an incremental setting.
\newblock \emph{arXiv preprint arXiv:2302.12049}.

\bibitem[{Williams et~al.(2018)Williams, Kannan, Aleksic, Rybach, and Sainath}]{williams2018contextual}
Ian Williams, Anjuli Kannan, Petar~S Aleksic, David Rybach, and Tara~N Sainath. 2018.
\newblock Contextual speech recognition in end-to-end neural network systems using beam search.
\newblock In \emph{Interspeech}, pages 2227--2231.

\bibitem[{Zayyan(2011)}]{zayyan2011objective}
Marliyya Zayyan. 2011.
\newblock Objective structured clinical examination: the assessment of choice.
\newblock \emph{Oman medical journal}, 26(4):219.

\bibitem[{Zhang and Zhou(2022)}]{zhang2022end}
Zhengyi Zhang and Pan Zhou. 2022.
\newblock End-to-end contextual asr based on posterior distribution adaptation for hybrid ctc/attention system.
\newblock \emph{arXiv preprint arXiv:2202.09003}.

\bibitem[{Zhao et~al.(2019)Zhao, Sainath, Rybach, Rondon, Bhatia, Li, and Pang}]{zhao2019shallow}
Ding Zhao, Tara~N Sainath, David Rybach, Pat Rondon, Deepti Bhatia, Bo~Li, and Ruoming Pang. 2019.
\newblock Shallow-fusion end-to-end contextual biasing.
\newblock In \emph{Interspeech}, pages 1418--1422.

\bibitem[{Zhao et~al.(2023)Zhao, Wang, and Wang}]{zhao2023knowledge}
Zihan Zhao, Yu~Wang, and Yanfeng Wang. 2023.
\newblock Knowledge-aware bayesian co-attention for multimodal emotion recognition.
\newblock In \emph{ICASSP 2023-2023 IEEE International Conference on Acoustics, Speech and Signal Processing (ICASSP)}, pages 1--5. IEEE.

\bibitem[{Zhong et~al.(2019)Zhong, Wang, and Miao}]{zhong2019knowledge}
Peixiang Zhong, Di~Wang, and Chunyan Miao. 2019.
\newblock Knowledge-enriched transformer for emotion detection in textual conversations.
\newblock In \emph{Proceedings of the 2019 Conference on Empirical Methods in Natural Language Processing and the 9th International Joint Conference on Natural Language Processing (EMNLP-IJCNLP)}, pages 165--176.

\end{thebibliography}


\end{document}